\pdfoutput=1
\documentclass[11pt]{article}

\usepackage[]{acl}

\usepackage{times}
\usepackage{latexsym}
\usepackage{enumerate}
\usepackage{mathtools}
\usepackage{amsmath}
\usepackage{graphicx} %插入图片的宏包
\usepackage{float} %设置图片浮动位置的宏包
\usepackage{subfigure} %插入多图时用子图显示的宏包
\usepackage{color}
\usepackage{hyperref}
\usepackage{multicol}
\usepackage{booktabs}
\usepackage{tabularray}
\usepackage{multirow}
\usepackage{stfloats}
\usepackage{tablefootnote}
\hypersetup{hidelinks, colorlinks=true, 
        citecolor=darkblue,
        urlcolor=darkblue,
        pdfstartview=Fit, 
        breaklinks=true}

\definecolor{dark}{rgb}{195,89,89}
\usepackage[T1]{fontenc}
\usepackage[utf8]{inputenc}

\usepackage{microtype}

\title{BNS-Net: A Dual-channel Sarcasm Detection Method Considering Behavior-level and Sentence-level Conflicts}

\author{Liming Zhou\textsuperscript{1}, Xiaowei Xu\textsuperscript{*,2}, Xiaodong Wang\textsuperscript{3}}

\begin{document}
\maketitle
\begin{abstract}
Sarcasm detection is a binary classification task that aims to determine whether a given utterance is sarcastic. Over the past decade, sarcasm detection has evolved from classical pattern recognition to deep learning approaches, where features such as user profile, punctuation and sentiment words have been commonly employed for sarcasm detection.  In real-life sarcastic expressions, behaviors without explicit sentimental cues often serve as carriers of implicit sentimental meanings. Motivated by this observation, we proposed a dual-channel sarcasm detection model named BNS-Net. The model considers behavior and sentence conflicts in two channels. Channel 1: Behavior-level Conflict Channel reconstructs the text based on core verbs while leveraging the modified attention mechanism to highlight conflict information. Channel 2: Sentence-level Conflict Channel introduces external sentiment knowledge to segment the text into explicit and implicit sentences, capturing conflicts between them. To validate the effectiveness of BNS-Net, several comparative and ablation experiments are conducted on three public sarcasm datasets. The analysis and evaluation of experimental results demonstrate that the BNS-Net effectively identifies sarcasm in text and achieves the state-of-the-art performance.
\end{abstract}

\section{Introduction}
 Language is far from being a simple concatenation of words adhering to grammatical rules, it is inherently intertwined with the expression of emotions. \citet{camp2012sarcasm} contends that when people speak, they convey not only surface meanings but also implicit connotations. These implicit connotations, known as implicatures, can be revealed through reasoning. However, the presence of sarcasm hinders traditional implicature analysis methods from fully explaining the meanings in linguistic behavior. In other words, genuine sentiment in language is often obscured, making it difficult for traditional sentiment analysis methods to accurately identify the underlying sentiment.

Due to the complex and ambiguous linguistic features of sarcasm, identifying sentiment conflicts has become exceedingly challenging. Despite the fact that current deep learning models have reduced the cost of feature engineering to some extent, their interpretability and generalizability remain relatively low. Additionally, our study of real-world samples has revealed that some sarcastic texts do not have obvious sentiment conflicts. The inconsistency in context is not expressed through the opposition of sentiment words.

To address the above problems, we propose a sarcasm detection model called BNS-Net, the model contains two channels: behavior-level conflict channel and sentence-level conflict channel. The behavior-level conflict channel takes the verb and auxiliary verb in the text as the core to reconstruct the text, and focuses on the behavior conflict information in the sentence by using the Conflict Attention Mechanism(CAM). The sentence-level conflict channel introduces the external emotional knowledge, and splits the text into explicit semantic and implicit semantic sentences to capture the sentence-level conflict between the two sentences. Sentence-level conflict channel is to segment the text into explicit semantic and implicit semantic sentences to capture the sentence-level conflict between them. Experimental results on three public datasets show that BNS-Net has significantly improved performance compared to previous work, especially on the short text dataset.

The main contributions of this work are summarized as follows:
\begin{enumerate}[(1)]
    \item We proposed a novel sarcasm detection model named BNS-Net. To the best of our knowledge, this study is the first attempt to utilize the behavior-level conflict to construct contextual inconsistencies.
    \item  The Conflict Attention Mechanism we proposed improves the model's ability to interact and mine conflicts in the text, leading to more accurate and robust sarcasm detection.
    \item  The experimental results on three datasets demonstrate that the proposed BNS-Net model achieves state-of-the-art performance.
\end{enumerate}

\section{Related Work}
Sarcasm detection is commonly regarded as a binary classification problem, aiming to determine whether the target text contains sarcasm. Similar to other tasks in NLP, the development of sarcasm detection can be divided into three stages: (1) rule-based methods, (2) statistical learning-based methods, and (3) deep learning-based methods.

\subsection{Rule-based Methods}
 Earlier sarcasm detection tasks primarily used rule-based methods, relying on the detection of specific syntactic structures or salient constituents. For instance, \citet{veale2010detecting} used a specific query pattern "as * as a *" to find explicit sarcastic sentences on Google. \citet{riloff2013sarcasm} proposed an sentiment conflict-based sarcasm detection method, which identified positive and negative scenarios in sentences and explained sarcasm formation through conflict comparison, and this sentiment conflict-based method has been confirmed to be effective by several experiments. Additionally, \citet{bharti2015parsing} 
 proposed a sarcasm detection method that not only identified sentiment conflicts in the text but also considered the use of mood words and intensifiers. 
 
 However, these rule-based methods are not ideal as they heavily rely on sentiment lexicons and linguistic experts.

\subsection{Statistical Learning-based Methods}
 statistical-based methods are mainly applied during the middle stage of sarcasm detection research. These methods leverage statistical learning techniques to identify patterns and relationships within the data, enabling the model to make predictions based on learned statistical properties.
 
During this stage, researchers explored various statistical features and algorithms to improve the performance of sarcasm detection. \citet{gonzalez2011identifying} focused on user comments, emoticons, sentiment lexicons, and other features to classify sarcasm in textual data, they employed Support Vector Machine \cite{cortes1995support} along with Sequential Minimal Optimization \cite{platt1998sequential} and Logistic Regression\cite{cox1958regression} methods. \citet{reyes2012humor} utilized features like ambiguity, unexpectedness, and sentiment contexts. The semantic relevance was measured by using unexpectedness values to identify whether the target text contains sarcasm. \citet{joshi2015harnessing} employed features related to pragmatics, vocabulary, explicit and implicit inconsistencies, and combined two types of inconsistencies to detect sarcasm using the LibSVM classifier with an RBF kernel. \citet{reyes2012making} treated sarcasm as subjective language and proposed a model based on six feature categories and conducted experiments with three classifiers such as Decision Trees, SVM, and Naive Bayes. Additionally, \citet{ghosh2015sarcastic} analyzed and explained sarcasm formation from the perspective of word sense disambiguation (i.e., determining whether the meaning of the target word in the sentence is literal or sarcastic) and used an SVM classifier with the MVME kernel for sarcasm detection.

\subsection{deep learning-based methods}
Starting from a decade ago, there has been a gradual inclination towards employing deep learning-based methods for sarcasm detection. These methods leverage the power of deep learning techniques to automatically learn and extract complex features from the data.

As the field of deep learning continues to evolve, researchers are likely to explore more sophisticated architectures and advanced techniques to further enhance the effectiveness and efficiency of sarcasm detection. \citet{amir2016modelling} applied convolutional operations to generate user embeddings and text embeddings by learning from user's historical posting data, which were jointly used as critical information for sarcasm detection. \citet{hazarika2018cascade} proposed the CASCADE model, which leverages users' historical comments to encode their writing style and personality traits. \citet{ghosh2016fracking} proposed a multi-network fused deep neural network model. In this model, a CNN network with sigmoid activation function is used to extract discriminative word sequences as combined features for the LSTM\cite{hochreiter1997long}. \citet{ghosh2018sarcasm} employed multiple types of LSTM networks to model the context and showed that attention-based LSTM outperformed other networks on both long and short text datasets. \citet{xu2020reasoning} proposed a model for extracting differences and semantic correlations between modalities by constructing a Decomposition and Relation network (D\&R network) to model cross-modal contrasts and semantic associations. \citet{tay2018reasoning} were the first to introduce the intra attention mechanism for sarcasm detection. 

 Some of features are helpful in task, such as, Polarity Contrast\cite{van2017can}, Common Sense Knowledge\cite{van2018we}, Sentimental Features\cite{farias2016irony}, Negation\cite{giora1995irony} and Contextual Differences\cite{wilson2006pragmatics}. \citet{chen2022jointly} combined sentiment and conflict inconsistency features to model the target text. \citet{zhang2022novel} proposed a retrospective reading Chinese sarcasm detection model, which consists of two parallel modules: Sketchy Reading and Intensive Reading. In addition, masked and generative sarcasm detection methods have also emerged. \citet{wang2022masking} proposed a method that involves masking important words in the original text, they utilized BERT\cite{devlin2018bert} to encode the sentences with the masked parts filled using BART\cite{lewis2019bart}.\citet{liu2022dual} proposed a novel approach for sarcasm detection based on the contradictory nature of sarcasm. They utilized existing sentiment lexicon and applied rules to split sentences according to the frequency of sentiment words. In recent years, pre-trained models based on large unlabeled corpora have shown excellent performance in various NLP tasks. \citet{wang2021performance} evaluated several representative pre-trained models in sarcasm detection task and found that these PLMs are susceptible to the influence of the corpus and may make biased judgments due to preconceived notions. 

 However, the above methods mainly consider single-word or sentence-level granularity, and they do not fully take into account the role of phrases in conveying conflict information. Even with large-scale data and pretrained models, they still lack a deep understanding of the context of sarcasm. Considering the current trend of research and the investigation of datasets, we recognize the importance of behaviors and sentences in conveying inconsistency. Hence, we propose a dual-channel sarcasm detection approach that incorporates external emotional knowledge and simultaneously extracts conflict inconsistency information from both the "behavior-level" and "sentence-level" perspectives.

\section{Method}
In this chapter, we present a detailed exposition of the BNS-Net. The model consists of two channels: The Behavior-level Conflict Channel and the Sentence-level Conflict Channel. In the behavior-level conflict channel, we reconstruct the text with the verbs in the text as the core, and utilize the internal interaction ability of the CAM to highlight the behavior-level conflict information within texts. In the sentence-level conflict channel, we introduce external sentimental knowledge to segment the text into explicit semantic and implicit semantic sentences. The architecture of the model is illustrated in Figure 1.

\begin{figure}
    \centering
    \includegraphics[width=1\linewidth]{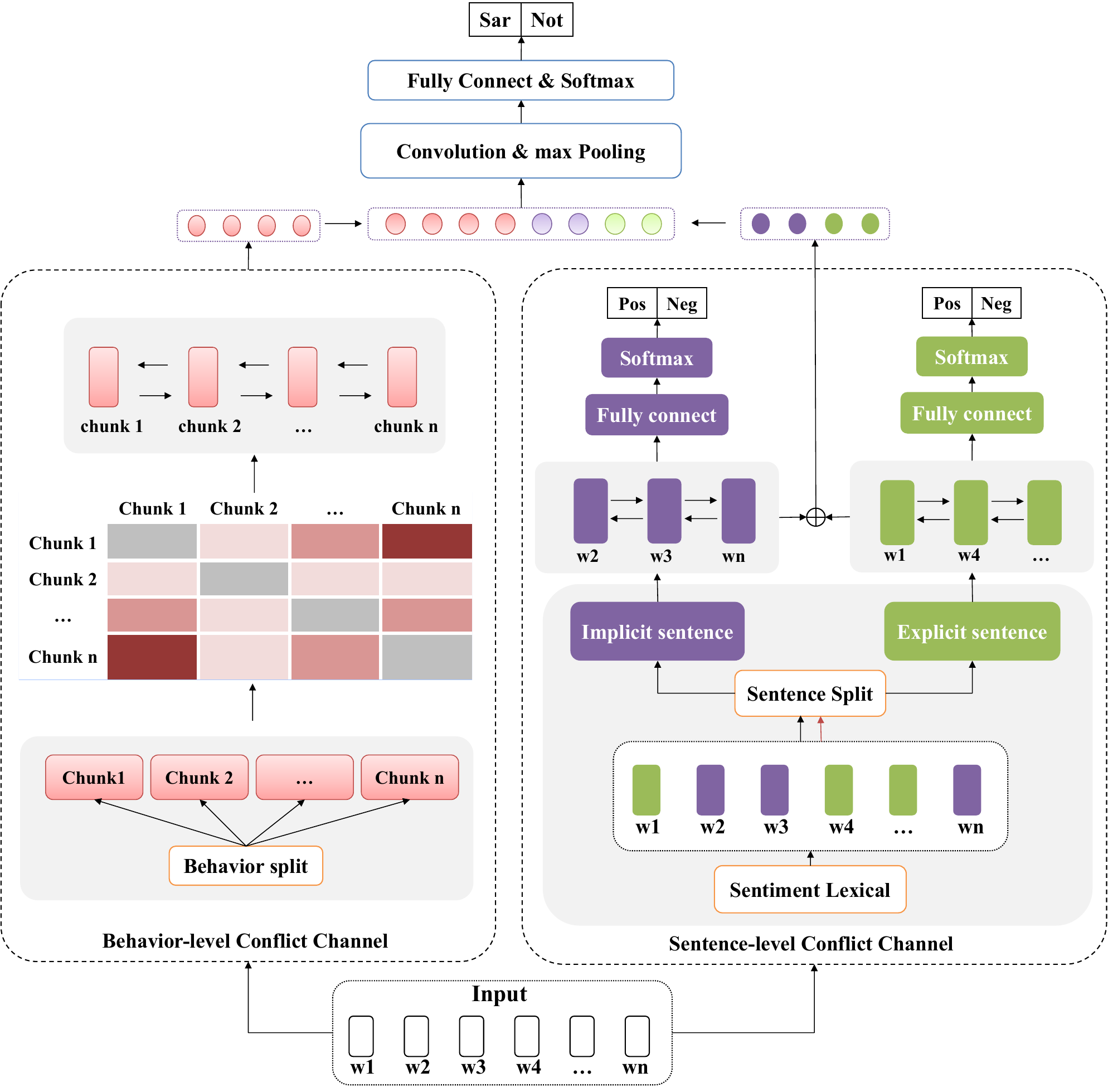}
    \caption{Overview of our proposed model (BNS-Net).}
    \label{fig:enter-label}
\end{figure}
The proposed model consists of several steps: Firstly, the input is mapped to the vector space. Next, the mapped input is simultaneously fed into two channels for sarcasm information extraction. The behavior-level conflict representations and sentence-level conflict representations output from these two channels respectively are then stacked together. Finally, a convolutional network with multi-scale kernels is used to further extract salient features to determine whether the target text is sarcastic. In the following sections, we will provide a detailed exposition of the two channels.

\subsection{Behavior-level Conflict Channel}

 To effectively highlight the conflict information in text, we designed the Behavior-level Conflict Channel, which incorporates bidirectional LSTM(Bi-LSTM) and a modified attention mechanism called Conflict Attention Mechanism(CAM). This mechanism is inspired by the human brain's ability to detect and resolve conflicts in stimulus and response processing.

This channel is mainly divided into three steps. Firstly, it employs behavior sliding segmentation to reconstruct the origin text. Next, it focuses on the parts of the text containing behavior conflicts through conflict attention mechanism. Finally, it uses Bi-LSTM to generate behavior-level conflict representations. The network structure of this channel is illustrated on the left side of Figure 1. The specifics of each step are detailed below.

\subsubsection{Behavior Sliding Segmentation}

To specifically extract the behavior elements from the text, we utilize the Spacy\footnote{\href{https://spacy.io/}{https://spacy.io/}} to perform parts-of-speech (POS) tagging on the text word by word before mapping it to the embedding layer. Spacy categorizes words into 17 classes, including verb, noun, auxiliary verb, adverb and more. Among these, verb and auxiliary verb are often associated with behaviors. Thus, we select words belonging to these two categories as centers and divide the original text into several chunks within a certain window size, which we called “behavior chunk”. The concept of sliding segmentation is illustrated in Figure 2.
\begin{figure}[H]
    \centering
    \includegraphics[width=7cm]{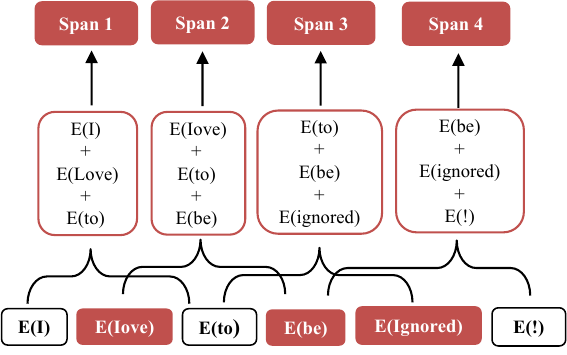}
    \caption{Schematic Diagram of Behavior Sliding Segmentation, where E(.)  represents the embedding corresponding to the token in the brackets}
    \label{fig:enter-label}
\end{figure}

The sentence “I love to be ignored!” is taken as an example in Figure 2. In this example, the core words are “love”, “be” and “ignored”. The inputs are word vectors mapped by the embedding layer, and a sliding window of size 3 is used to segment the sentence, resulting in segmented chunks i.e., \{I love to\}, \{love to be\}, \{to be ignored\} and \{be ignored!\}. Next, the embeddings of each token within each chunk are summed to obtain a vector called "behavior embedding", which contains the semantic information of each sub-vector in the current chunk. 

Since we base the sliding segmentation on two POS, “auxiliary verb” and “verb”, cases may arise where an auxiliary verb is followed by a verb. This situation can lead to the  repetitive behavior chunks, increasing data redundancy. To address this issue, the duplicated behavior chunks are eliminated firstly, next, the top $n$  most representative chunks corresponding to each core word are selected as its final behavior chunks.

\begin{equation}
n=\left \lceil \frac{window\ size}{2}  \right \rceil 
\end{equation}

In the above equation, $\left \lceil  \right \rceil$ represents the ceiling function. For chunk selection, we first conduct sentiment analysis on each token in the sentence using the SenticNet\footnote{\href{http://sentic.net/}{http://sentic.net/}}\cite{cambria2022senticnet} to obtain the sentiment intensity score of each token. Tokens without sentiment polarity are assigned 0, while tokens with negative sentiment are assigned the absolute value. Then, we select the top $n$ chunks for the current core word based on the aggregate sentiment intensity scores of all tokens in each chunk, as its final behavior chunks. 

\subsubsection{Behavior-level Conflict Focus}

After obtaining behavior embeddings, we focus on highlighting conflict between these chunks. Many previous studies have adopted attention mechanism and variant such as self-attention\cite{vaswani2017attention} and intra-ATTention\cite{tay2018reasoning} to emphasize text parts that are critical for sarcasm detection. As an encoding method to resolve long-distance dependencies, the core idea of attention is measuring the relevance and affinity between the Query and Key vectors via dot product (the more congruent the semantics, the higher the affinity).The formulas of self-attention mechanism are as follows.

\begin{equation}
    Q=X_{in}\cdot W_{q}
\end{equation}
\begin{equation}
    K=X_{in}\cdot W_{k}
\end{equation}
\begin{equation}
    V=X_{in}\cdot W_{v}
\end{equation}
\begin{equation}
    Out_{att}=softmax(\frac{Q \cdot K^{\tau } }{\sqrt{d_{k}}}) \cdot V 
\end{equation}

In the above equations, $X_{in}$  represents the input to the attention layer, $Q$, $K$ and $V$ are the query, key and value vectors obtained by multiplying the input vectors with their respective weight matrices(i.e., $W_{q}$, $W_{k}$, $W_{v}$ ), $d_{k}$ represents the dimension of the $K$, $Out_{att}$  represents the output of the attention mechanism. 

However, the above attention mechanism still has certain limitation for the sarcasm detection. It cannot guarantee that the observed key positions in the text precisely correspond to the ones responsible for sarcasm, which means that the idea of higher attention weights for tokens with higher similarity contradicts the principle of conflict inconsistency for sarcasm formation. We contend that the conflicting and inconsistent parts in sarcastic sentences should be semantically dissimilar, which is reflected at the word vector level as relatively lower dot product values between conflicting tokens.

In order to better highlight the significance of behavior-level conflicts in sarcasm detection, modification was made to the raw attention mechanism. This enhanced attention mechanism is referred to as the Conflict Attention Mechanism (CAM). Unlike raw attention mechanism, the softmin function is used to normalize the dot product results of vectors. The specific formula of CAM is shown below.

\begin{equation}
    softmin(x_i)=\frac{e^{-x_i}}{ \sum_{i=0}^{N}e^{-x_{i}}  } 
\end{equation}
\begin{equation}
    Out_{att}=softmin(\frac{Q \cdot K^{\tau } }{\sqrt{d_{k}}}) \cdot V 
\end{equation}

 As deduced from equation (6), the softmin function can scale n-dimensional data to the   interval and assign larger values (interpreted as weights) to smaller elements, which is opposite to softmax. The feasibility of this approach is supported by the Word2vec\cite{mikolov2013efficient}, a word embedding technique. When overlaps exist between some behavior chunks, their vector values become more similar, in other words, the dot product of them tends to be larger than that of two behavior chunks with fewer or no overlaps.

\subsubsection{Temporal Logic Encoding}
By modeling the behavior embeddings through the behavior-level conflict focus step, long-range dependency interactions between behavior chunks are achieved, and more attention is directed towards conflicting chunks. However, relying solely on the CAM, which is agnostic to chunk distances, would cause its output to lack sequential logic semantics between contexts. Therefore, the bidirectional LSTM(Bi-LSTM) models is introduced to endow behavior embeddings with additional sequential semantics.

As an excellent temporal model, Bi-LSTM takes into account the influence of temporal sequence information on semantic encoding and models the text from both front and back directions simultaneously. The specific equations are shown below, and for the sake of brevity, this paper refrains from elaborating on the detailed principles and equations within Bi-LSTM.

\begin{equation}
    \overrightarrow{h}_{i}^{t}=\overrightarrow{LSTM}(x_{i}^{t},\overrightarrow{h}_{i-1}^{t})       
\end{equation}
\begin{equation}
    \overleftarrow{h}_{i}^{t}=\overleftarrow{LSTM}(x_{i}^{t},\overleftarrow{h}_{i-1}^{t})       
\end{equation}
\begin{equation}
    h_{i}^{t}=[\overrightarrow{h}_{i}^{t};\overleftarrow{h}_{i}^{t} ]   
\end{equation}

In the above equations, $x_{i}^{t}$ represents each token in the sentence, $\overrightarrow{h}_{i}^{t}$ and $\overleftarrow{h}_{i}^{t}$ represent the forward and backward hidden states in the Bi-LSTM layer, respectively. We use the hidden state vector at the last time step $h$ as the representation of behavior-level conflicts formed in this channel.

\subsection{Sentence-level Conflict Channel}
The ambiguity of sarcasm can present a sentence with two different emotional meanings, making the intention of the sentence ambiguous. Based on this trait, we construct two sentences with different sentimental semantics based on the original text: the Implicit Semantic Sentence and the Explicit Semantic Sentence, to uncover textual incongruity at the sentence level. This channel consists of two steps: (1) Text Reconstruction and (2) Sentiment Classification. The network structure for this channel is illustrated on the right side of Figure 1.

\subsubsection{Text Reconstruction}

The SenticNet is employed to reconstruct the original text, thereby introducing external sentiment knowledge into the model. As the first step, we utilize SenticNet to perform sentiment analysis on each word and acquire the sentiment polarity and intensity scores of them. Subsequently, we select higher value between the positive and negative polarity as the surface-level sentiment of the original text. Furthermore, once the surface-level sentiment is determined, all parts of the sentence that align with the surface-level sentiment are extracted and form the explicit semantic sentence, while the remaining parts become the implicit semantic sentence. Figure 3 provides an illustrative representation of the sentence reconstruction process.
\begin{figure}[H]
    \centering
    \includegraphics[width=7cm]{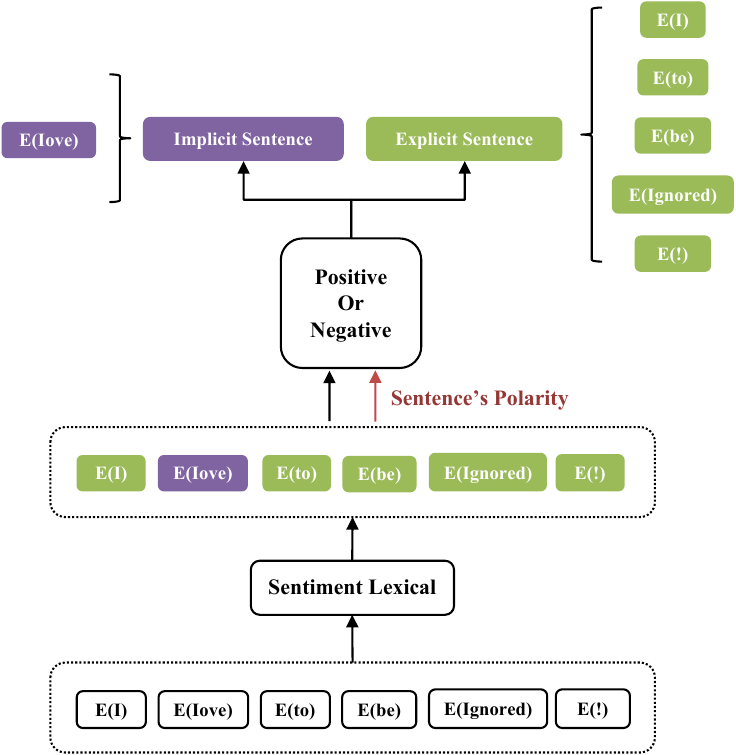}
    \caption{Schematic diagram of sentence reconstruction.}
    \label{fig:enter-label}
\end{figure}
To clearly illustrate the process of sentence reconstruction, the text "I love to be ignored!" is taken as an example (shown in Figure 3). In this example, because the positive sentiment polarity of \{love\} is stronger than the negative polarity of \{ignored\}. Therefore, the surface sentiment of this sentence is positive, and all the positive words are extracted from the original text as explicit semantic sentences, while the rest form implicit semantic sentences.

\subsubsection{Sentiment Classification}

After obtaining the explicit sentence and implicit sentence, the Bi-LSTM layer is employed to encode these two sentences. Following the Bi-LSTM encoding, we utilize fully-connected and softmax layers to perform sentiment classification on two sentences. Here two subtasks are introduced, sentiment classification on explicit and implicit sentences, and the cross-entropy function is used for computing their loss.
\begin{normalsize}
    \begin{equation}
    J = -\sum_{i=1}^{N}y_{i}\log \hat{y}_{i}+(1-y_{i})\log(1-\hat{y}_{i})   
\end{equation}
\end{normalsize}
In the above equation, $y_{i}$ represents the true label, $\hat{y}_{i}$ represents the predicted label.

\subsection{Training Objective}
The outputs of the behavior-level conflict channel and the sentence-level conflict channel are stacked together and fed into a convolutional network with multi-scale kernels, and the outputs of the fully-connected and softmax layers are used to determine whether the target text is sarcastic or not, where we introduce a third training sub-task, sarcastic classification, which is also optimized using the cross-entropy loss function. 

Since three subtasks are jointly trained, the final loss of this model is composed of the sentiment classification loss for explicit and implicit sentences, as well as the sarcasm classification loss. The specific formula is shown below.
\begin{equation}
    \mathcal{L} = \lambda_{1}\cdot J_{sar} + \lambda_{2}\cdot J_{imp}+ \lambda_{3}\cdot J_{exp}
\end{equation}
In equation (12), $\lambda_{1}$, $\lambda_{2}$ and $\lambda_{3}$ are utilized to balance the significance of different subtasks. $J_{sar}$, $J_{imp}$, $J_{exp}$ represent the losses of the three subtasks respectively. By minimizing the loss functions of all subtasks, the model gradually enhances its semantic encoding capability for both explicit and implicit sentences, ultimately improving the sarcasm detection ability of the model.

\section{Experiments}
This section aims to describe the details and results of the experiments. In this chapter, we will demonstrate the advantages of the proposed model with solid experimental support, discuss its interpretability and applicability, and understand the contribution and necessity of each module. 

\subsection{Datasets}
Experiments were conducted on three sarcasm datasets: IAC-V1, IAC-V2, and Twitter.
\begin{itemize}
    \setlength{\itemsep}{2pt}
    \setlength{\parsep}{2pt}
    \setlength{\parskip}{2pt}
    \item The \textbf{IAC-V1} dataset\cite{lukin2017really} is derived from the 4Forum, primarily used for political debates and voting. 
    \item The \textbf{IAC-V2} dataset\cite{oraby2017creating} is an upgraded version of IAC-V1, containing a larger amount of data. 
    \item The \textbf{Twitter} dataset is derived from Subtask A of the SemEval-2018 Task 3\cite{van2018semeval}. The data was filtered based on the presence of specific tags (\verb|#sarcasm|, \verb|#irony|, \verb|#not|) in the tweets. 
\end{itemize}
\begin{table}[t]
    \centering
    \caption{Statistics of dataset used in our experiments, where Sarc denotes the proportion of sarcastic texts, Ratio denotes the proportion of sarcastic texts without apparent conflicts among all sarcastic texts}
    \resizebox{\linewidth}{!}{
    \begin{tabular}{c|ccccc}
    \hline
    \textbf{Dataset} & \textbf{Train} & \textbf{Valid} & \textbf{Test} & \textbf{Sar(\%)} & \textbf{Ratio(\%)} \\
    \hline
    IAC-V1\tablefootnote{\href{https://nlds.soe.ucsc.edu/sarcasm1}{https://nlds.soe.ucsc.edu/sarcasm1}} & 1596 & 80 & 320 & 50.36 & 28.32 \\ 
    IAC-V2\tablefootnote{\href{https://nlds.soe.ucsc.edu/sarcasm2}{https://nlds.soe.ucsc.edu/sarcasm2}} & 5216 & 262 & 1042 & 50.01 & 23.91 \\ 
    Twitter\tablefootnote{\href{https://github.com/Cyvhee/SemEval2018-Task3}{https://github.com/Cyvhee/SemEval2018-Task3}} & 3526 & 100 & 784 & 51.18 & 51.89 \\ 
    \hline
    \end{tabular}
    }
    \label{tab:my_label}
\end{table}

 These datasets can be categorized into two types: one is long-text data (IAC-V1, IAC-V2) with text lengths ranging from 50 to 150, and the other is short-text data (Twitter) with text lengths ranging from 5 to 40. By observing the "Ratio" values, we found that a considerable portion of sarcastic texts in the existing corpus lack obvious sentiment conflicts, even reached about half of the proportion in Twitter dataset, indirectly validating the significance of our research. 

\subsection{Baseline Models}
\begin{itemize}
    \setlength{\itemsep}{2pt}
    \setlength{\parsep}{2pt}
    \setlength{\parskip}{2pt}
    \item Bi-LSTM\cite{hochreiter1997long} is a variant of LSTM that adds a backward LSTM layer to the traditional LSTM, allowing it to learn both forward and backward long-range dependency information simultaneously.
    \item Att-LSTM\cite{wang2016attention} is a network model that combines LSTM and attention mechanism. 
    \item CNN-LSTM-DNN\cite{ghosh2016fracking} is an approach that combines CNN, LSTM, and DNN networks. 
    \item SIARN\cite{tay2018reasoning} is a dual-channel network that utilizes the intra-ATTention mechanism. 
    \item MIARN\cite{tay2018reasoning} is a variant of SIARN that extends the single-dimensional intra-ATTention computation of SIARN to multiple dimensions.
    \item DC-Net\cite{liu2022dual} is a dual-channel sarcasm detection model that reconstructs the text into literal semantics and implied semantics separately and models them independently.
\end{itemize}
The aforementioned models are solely considering the target text without taking into account additional information. This aligns with the direction of our research.
\subsection{Implementation Details}
For experiments, the datasets were preprocessed and all the models were evaluated using evaluation metrics such as Precision, Recall, Macro-F1, and Accuracy. The experiments were conducted on the Linux system, utilizing four Nvidia 2080ti GPUs and the PyTorch 2.0 framework for model training. For fair comparison, the embedding layers of all models were initialized with a 300-dimensional word2vec model trained on the Google News corpus. The batch size for training was set to 32 for all models. 

 For BNS-Net, the AdamW\cite{loshchilov2018fixing} optimizer was utilized for model training. On the Twitter dataset, the model's learning rate was set to 1e-4. On the IAC dataset, the model's learning rate was set to 2e-4. The dropout ratio was set to 0.5, L2 regularization was set to 0.01, ReLU was used as the activation function, attention heads were set to 10, and the Bi-LSTM had 2 layers. 

% \begin{table*}[]
%     \centering
%     \renewcommand{\arraystretch}{1.5}
%     \begin{tabular}{ccccccccccccccc}
%     \hline
%     \multirow{2}{*}{Model}  &  \multicolumn{4}{c}{IAC-V1}  & \ & \multicolumn{4}{c}{IAC-V2} &\ &  \multicolumn{4}{c}{Twitter-2018} \\
%     \cmidrule{2-5} \cmidrule{7-10} \cmidrule{12-15}
%     \ & Pre. & Rec. & F1 & Acc. & \ & Pre. & Rec. & F1 & Acc. & \ & Pre. & Rec. & F1 & Acc.\\
%     \hline
    
%          & 
%     \end{tabular}
    
%     \caption{Caption}
%     \label{tab:my_label}
% \end{table*}

\begin{table*}[]
\centering
\caption{Performance of different sarcasm detection models on IAC-V1, IAC-V2, Twitter datasets, Where Pre denotes Precision, Rec denotes Recall, F1 denotes Macro-F1, Acc denotes Accuracy.}
\renewcommand{\arraystretch}{1.2}
\resizebox{\linewidth}{!}{
\begin{tabular}{ccccccccccccccc}
\hline
\multirow{2}{*}{Model} & \multicolumn{4}{c}{IAC-V1} &  & \multicolumn{4}{c}{IAC-V2} &  & \multicolumn{4}{c}{Twitter} \\ \cline{2-5} \cline{7-10} \cline{12-15} 
 & Pre. & Rec. & F1 & Acc. &  & Pre. & Rec. & F1 & Acc. &  & Pre. & Rec. & F1 & Acc. \\ \hline
Vanilla Bi-LSTM & 58.78 & 58.19 & 58.39 & 58.47 &  & 61.87 & 61.87 & 61.87 & 61.88 &  & 58.16 & 57.88 & 57.51 & 57.88 \\
ATT-LSTM & 63.11 & 61.41 & 60.06 & 61.34 &  & 63.73 & 63.47 & 63.59 & 63.47 &  & 62.92 & 62.86 & 62.86 & 62.86 \\
CNN-LSTM-DNN & 59.93 & 59.72 & 59.53 & 59.74 &  & 64.85 & 64.79 & 64.81 & 64.78 &  & 63.22 & 63.18 & 63.16 & 63.18 \\
SIARN & 66.24 & 59.01 & 53.92 & 59.11 &  & 71.37 & 66.75 & 64.84 & 66.73 &  & 60.35 & 59.49 & 58.62 & 59.49 \\
MIARN & 61.14 & 59.69 & 58.36 & 59.74 &  & 70.91 & 68.38 & 67.38 & 68.36 &  & 64.01 & 63.18 & 62.64 & 63.18 \\
DC-Net & 62.93 & 62.33 & 61.88 & 62.31 &  & 75.26 & 74.87 & 74.78 & 74.88 &  & 68.61 & 68.49 & 68.44 & 68.49 \\
\textbf{BNS-Net(ours)} & 66.53 & 66.16 & 65.95 & 66.13 &  & 75.98 & 75.93 & 75.92 & 75.93 &  & 73.58 & 73.47 & 73.44 & 73.47 \\ \hline
\end{tabular}}
\end{table*}

\subsection{Analysis of Experimental Results}
Table 2 presents the performance comparison of the models on the Twitter and IAC datasets. It is observed that BNS-Net model outperforms previous models across all evaluation metrics, particularly on the Twitter and IAC-V1 datasets, where the metrics show a significant improvement of 4 to 5 percentage points. The improvement on the IAC-V2 dataset is relatively smaller, approximately 1 to 2 percentage points.

Regarding the insignificant improvement in our model's performance on the IAC-V2 dataset, we attribute it to the dataset's large volume, potentially containing more noise and redundancy. Long texts' intricate semantic complexity may also introduce challenging patterns. Increased behavior chunk generation during sliding segmentation contributes to data redundancy, causing network learning oscillations. While limiting behavior chunks reduces data volume, it may truncate important information. 

\begin{table}[t]
\centering
\caption{ Results of the ablation experiments, where del-S indicates that the sentence-level conflict channel is deleted, del-B indicates that the behavior-level conflict channel is deleted, raw-ATT indicates that the original attention mechanism is used, and no-subloss indicates that multi-task learning is not used.}
\renewcommand{\arraystretch}{1.2}
\resizebox{6.8cm}{!}{
\begin{tabular}{ccccc}
\hline
\multirow{2}{*}{Model} & \multicolumn{4}{c}{Twitter-2018} \\ \cline{2-5} 
 & Pre. & Rec. & F1 & Acc. \\ \hline
BNS(del-S) & 65.28 & 65.27 & 65.27 & 65.27 \\
BNS(del-B) & 69.76 & 69.22 & 69.04 & 69.22 \\
BNS(raw-ATT) & 71.12 & 70.74 & 70.61 & 70.74 \\
BNS(no-subloss) & 71.79 & 71.71 & 71.67 & 71.71 \\ \hline
BNS & 73.58 & 73.47 & 73.44 & 73.47 \\ \hline
\end{tabular}}
\end{table}
\subsection{Ablation Study}
In this chapter, a series of ablation experiments were conducted to assess the impact of different components in the model. A set of ablation experiments was designed on the Twitter dataset, and to ensure the effectiveness and scientific rigor of these experiments, other variables were fixed during the experiments. The experimental configurations are presented in Table 3. The results of the ablation experiments will be analyzed separately for each of the two channels.

\subsubsection{On Behavior-level Conflict Channel}
In our proposed model, the behavior-level conflict channel no longer merely models the conflict between words, but rather reconstructs the original text at the granularity of "behavior chunks" through sliding segmentation. The construction of behavior chunks and the CAM are vital components of this channel, so we hypothesize that both of them contribute significantly to the model's performance.
To validate our hypothesis, analysis was conducted from three aspects: (a) whether this channel is effective for sarcasm detection, (b) if effective, what window size is most suitable for sliding segmentation, (c) whether the CAM can capture conflict information.

Regarding aspect (a), the BNS (del-B) in Table 3 directly shows the results after deleting the behavior-level conflict channel: the model performance decreased, with around 4 percentage points drop across metrics. This indicates that this channel does make significant contributions. 

Regarding aspect (b), during the process of constructing behavioral chunks, different sliding window sizes lead to variations in the number of behavioral chunks obtained through sliding segmentation. To investigate the impact of window size on model performance, four sets of experiments were conducted with window sizes set to [2, 3, 4, 5].
\begin{figure}[t]
    \centering
    \includegraphics[width=7.5cm]{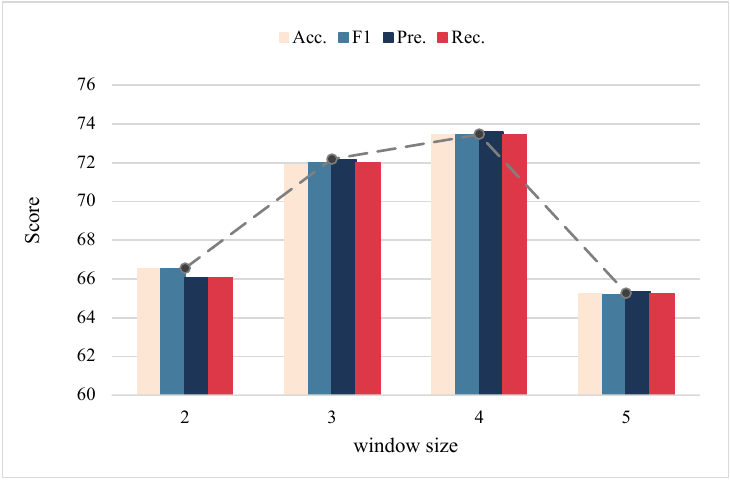}
    \caption{Comparison of evaluation metrics when the window size is 2, 3, 4 and 5 respectively.}
    \label{fig:enter-label}
\end{figure}

As observed in Figure 5, model performance shows an increasing trend as the window size grows until the size of 5. Optimal results across metrics are achieved at window size of 4. 

Regarding aspect(c), the BNS (raw-ATT) in Table 3 presents the results of replacing the CAM with the raw attention mechanism, which leads to a decrease of approximately 3\% in all evaluation metrics. Accordingly, we did further research.

% \usepackage{multirow}
\begin{table*}
\renewcommand{\arraystretch}{1.5}
\caption{Visualization of normalized attentional weights on three different attentional models, 
the intensity denotes the strength of the attention weight on the behavior blocks.}
\resizebox{\linewidth}{!}{
    \begin{tabular}{c|c}
\hline
\multirow{2}{*}{Model} & Input \\ \cline{2-2} 
 & Monsters like to take human brains, but trust me, you are safe. \\ \hline
raw-ATT & \colorbox[HTML]{C35959}{{[}Monsters like to{]}} \colorbox[HTML]{E7BBBE}{{[}like to take{]}} \colorbox[HTML]{F8E9E8}{{[}to take human{]}} \colorbox[HTML]{F8E9E8}{{[}take human brains{]}} \colorbox[HTML]{E7BBBE}{{[}barins but trust{]}} \colorbox[HTML]{C35959}{{[}but trust me{]}} \colorbox[HTML]{F8E9E8}{{[}me you are{]}} \colorbox[HTML]{E7BBBE}{{[}you are safe{]}} \\
intra-ATT & \colorbox[HTML]{C35959}{{[}Monsters like to{]}} \colorbox[HTML]{C35959}{{[}like to take{]}} \colorbox[HTML]{F8E9E8}{{[}to take human{]}} \colorbox[HTML]{F8E9E8}{{[}take human brains{]}} \colorbox[HTML]{E7BBBE}{{[}barins but trust{]}} \colorbox[HTML]{C35959}{{[}but trust me{]}} \colorbox[HTML]{F8E9E8}{{[}me you are{]}} \colorbox[HTML]{E7BBBE}{{[}you are safe{]}} \\
\textbf{CAM(ours)} & \colorbox[HTML]{C35959}{{[}Monsters like to{]}} \colorbox[HTML]{E7BBBE}{{[}like to take{]}} \colorbox[HTML]{F8E9E8}{{[}to take human{]}} \colorbox[HTML]{C35959}{{[}take human brains{]}} \colorbox[HTML]{E7BBBE}{{[}barins but trust{]}} \colorbox[HTML]{C35959}{{[}but trust me{]}} \colorbox[HTML]{F8E9E8}{{[}me you are{]}} \colorbox[HTML]{C35959}{{[}you are safe{]}} \\ \hline
\end{tabular}}
\end{table*}

Due to the small performance difference between window sizes 3 and 4 and for the sake of brevity, we chose window size 3 as an example. Table 4 presents a visual comparison of the three attention mechanisms. 

We observed that both the raw-ATT and the intra-ATT (from the MIARN) can focus on some important chunks in the text, but not necessarily on the conflicting chunks of speech that form the sarcasm. The attention given to conflicting chunks is relatively limited. This result is reasonable, because raw-ATT and the intra-ATT are designed to find affinities and similarities between word pairs, reallocating weights through the softmax function. However, this approach does not fully align with the context of sarcasm and conflict. In contrast, the CAM is able to identify the conflicting parts within the sentence, which exhibit minimal overlap. It utilizes the softmin function to focus on positions in the text with low semantic similarity and affinity, which coincides with the conflict and inconsistency principle of sarcasm detection and enables the network to generate attention representations which are more beneficial for sarcasm detection.

\subsubsection{On Sentence-level Conflict Channel}

Through analyzing the dataset, we found that the method of contrasting sentences by dividing them into explicit and implicit ones can partially explain the formation of sarcasm. Therefore, we tentatively believe that sentence-level conflict detection is also of significant part in sarcasm detection. 

To validate our hypothesis, analysis was conducted from three aspects: (a) whether this channel is effective for sarcasm detection, (b) if effective, whether this multi-task learning approach can improve model performance.

Regarding aspect (a), the BNS (del-S) in Table 3 directly shows the results after deleting the sentence-level conflict channel: the metrics dropped around 8\%, a large decrease. This outcome highlights the importance of sentence-level conflicts in sarcasm detection. 
\begin{figure}[t]
    \centering
    \includegraphics[width=7.5cm]{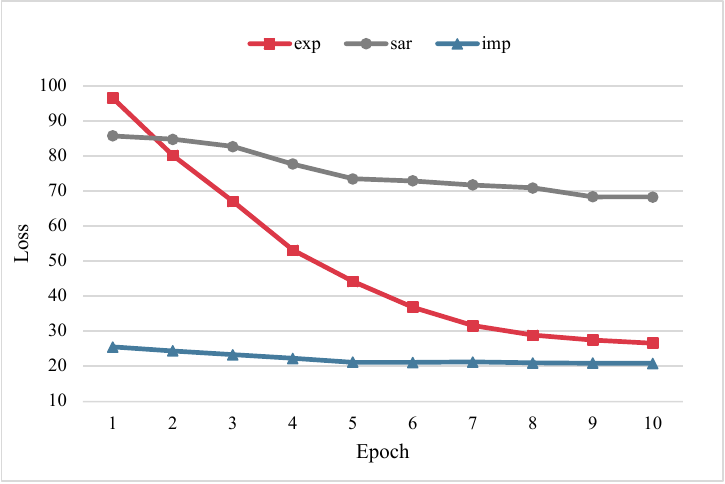}
    \caption{Loss curves for the subtasks.}
    \label{fig:enter-label}
\end{figure}

Regarding aspect (b), Table 3 presents the results of BNS (no-subloss), which denotes the model learning solely based on the sarcasm labels of texts. The model experienced a minor decrease of approximately 2\% in all evaluation metrics. This indicates that the current multi-task learning approach has a positive impact on the model, as it leverages the complementary nature between sarcasm detection and sentiment analysis tasks, integrating sentimental and sarcastic information, which mutually reinforce each other. 

Regarding aspect (c), the loss curves for three subtasks are shown in Figure 5. It can be observed that all three curves exhibit a decreasing trend. Among them, the sentiment classification task for explicit sentences demonstrates the most efficient learning progress, followed by the sarcasm classification task, while the sentiment classification task for implicit sentences shows the slowest learning progress.

The reason behind these observations lies in the fact that the explicit sentences contains obvious sentimental word components, making it relatively easier for sentiment classification. While, implicit sentences may not contain obvious sentimental words and even more noise. Comparatively, the sentiment analysis of implicit sentences is inherently a more challenging classification task than the other two tasks. 

\section{Conclusions}
In this paper, we propose sarcasm detection approach that considers behavior-level and sentence-level conflicts. The behavior-level conflict channel utilizes CAM to capture conflict information between behavior chunks, enabling the model to focus on conflicting parts in the text.The sentence-level conflict channel explains sarcasm formation from a more macroscopic level perspective. It leverages external knowledge to divide the sentence into explicit and implicit sentences for sentiment analysis. To validate our approach, experiments were conducted on three sarcasm datasets and comparisons were made with several baselines. Results demonstrate that our method achieves current state-of-the-art performance.

Although some progress has been achieved, there are still limitations. Future research efforts will focus on improving the methods for implicit sentiment mining and feature extraction. We expect that future studies will advance the development of deep learning in the field of sarcasm detection and provide more effective solutions to sentiment and sarcastic-related issues.

\section*{Acknowledgements}
This work was supported in part by the National Key Research and Development Program of China under Grant 2020YFB1710005; in part by Natural Science Foundation of Shandong Province under Grant ZR2022MF299.

\bibliography{anthology,custom}
\bibliographystyle{acl_natbib}

\end{document}